%% file: 0_main.tex
\documentclass[final]{cvpr}

\usepackage{times} 

\usepackage{epsfig}
\usepackage{graphicx}
\usepackage{amsmath}
\usepackage{amssymb}
\usepackage{booktabs}
\usepackage{multirow, xspace}
\usepackage{paralist}
\usepackage{algorithm}
\usepackage{algorithmic}
\usepackage{microtype}

\graphicspath{{./}{./figures/}}
\def\Vec#1{{\boldsymbol{#1}}}
\def\Mat#1{{\boldsymbol{#1}}}
\def\Ten#1{{\mathcal{#1}}}

\def\CDT{Cross-Domain Triplet } 

\usepackage[pagebackref=true,breaklinks=true,colorlinks,bookmarks=false,citecolor=blue,linkcolor=blue]{hyperref}

\def\cvprPaperID{10227} 
\def\confYear{CVPR 2021}
\begin{document}
\title{Cross-Domain Similarity Learning for Face Recognition in Unseen Domains}

\author{Masoud Faraki$^1$ ~~~~  Xiang Yu$^1$  ~~~~  Yi-Hsuan Tsai$^1$  ~~~~ Yumin Suh$^1$  ~~~~ Manmohan Chandraker$^{1,2}$\\ \\
$^1$NEC Labs America\\
$^2$University of California, San Diego\\
}

\maketitle

\begin{abstract}
Face recognition models trained under the assumption of identical training and test distributions often suffer from poor generalization when faced with unknown variations, such as a novel ethnicity or unpredictable individual make-ups during test time. In this paper, we introduce a novel cross-domain metric learning loss, which we dub \CDT (CDT) loss, to improve face recognition in unseen domains. The CDT loss encourages learning semantically meaningful features by enforcing compact feature clusters of identities from one domain, where the compactness is measured by underlying similarity metrics that belong to another training domain with different statistics. Intuitively, it discriminatively correlates explicit metrics derived from one domain, with triplet samples from another domain in a unified loss function to be minimized within a network, which leads to better alignment of the training domains. The network parameters are further enforced to learn generalized features under domain shift, in a model-agnostic learning pipeline. Unlike the recent work of Meta Face Recognition~\cite{guo2020learning}, our method does not require careful hard-pair sample mining and filtering strategy during training. Extensive experiments on various face recognition benchmarks show the superiority of our method in handling variations, compared to baseline and the state-of-the-art methods. 
\end{abstract}




\input{1_intro.tex}

\input{2_related.tex}

\input{3_method.tex}

\input{4_exp.tex}
\section{Conclusions}
We have introduced a cross-domain metric learning loss, which is dubbed \CDT (CDT) loss, that leverages the information jointly contained in two observed domains to provide better alignment of the domains. In essence, it first, takes into account similarity metrics of one data distribution, and then in a similar fashion to the triplet loss, uses the metrics to enforce compact feature clusters of identities that belong to another domain. Intuitively, CDT loss discriminatively correlates explicit metrics obtained from one domain with triplet samples from another domain in a unified loss function to be minimized within a network, which leads to better alignment of the training domains. We have also incorporated the loss in a meta-learning pipeline, to further enforce the network parameters to learn generalized features under domain shift. Extensive experiments on various face recognition benchmarks have shown the superiority of our method in handling variations, ethnicity being the most important one.  In the future, we will investigate how a universal form of covariance matrix (such as the one used in~\cite{chen2019progressive}) could be utilized in our framework. 


\clearpage
{\small
\bibliographystyle{ieee_fullname}
\bibliography{egbib}
}

\end{document}

%% file: 1_intro.tex
\vspace{-0.2cm}
\section{Introduction}
\vspace{-0.2cm}

Face recognition using deep neural networks has shown promising outcomes on popular evaluation benchmarks ~\cite{huang2008labeled, kemelmacher2016megaface, klare2015pushing, kalka2018ijb}. Many current methods base their approaches on the assumption that the training data -- CASIA-Webface~\cite{yi2014learning} or MS-Celeb-1M~\cite{guo2016ms} being the widely used ones -- and the testing data have similar distributions. However, when deployed to real-world scenarios, those models often do not generalize well to test data with unknown statistics. In face recognition applications, this may mean a shift in attributes such as ethnicity, gender or age between the training and evaluation data. On the other hand, collecting and labelling more data along the underrepresented attributes is costly. Therefore, given existing data, learning algorithms are needed to yield universal face representations and in turn, be applicable across such diverse scenarios. 

Domain generalization has recently emerged to address the same challenge, but mainly for object classification with limited number of classes ~\cite{balaji2018metareg,dou2019domain,li2019feature}. It aims to employ multiple labeled source domains with different distributions to learn a model that generalizes well to unseen target data at test time. However, many domain generalization methods are tailored to closed-set scenarios and hence, not directly applicable if the label spaces of the domains are disjoint. Generalized face recognition is indeed a prominent example of open-set applications with very large number of categories, encouraging the need for further research in this area. 

\begin{figure}[!tb]
\centering
\includegraphics[width=0.45\textwidth]{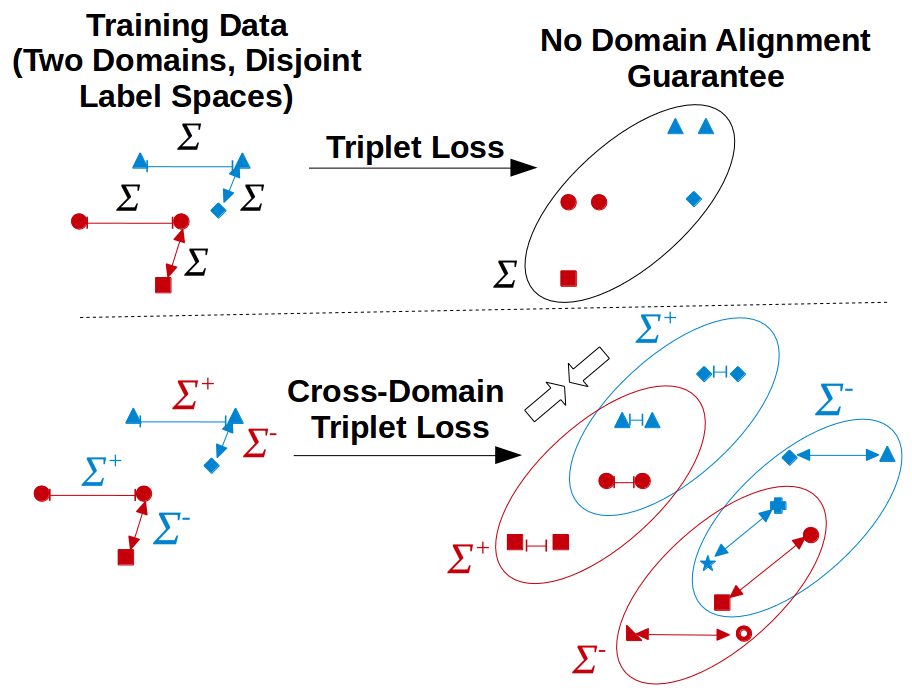}
\caption{Comparison between the conventional triplet and our \CDT losses. \textbf{Top:} The standard triplet loss is domain agnostic and utilizes a shared metric matrix, $\Mat{\Sigma}$, to measure distances of all (anchor,positive) and (anchor,negative) pairs.  \textbf{Bottom:} Our proposed \CDT loss, takes into account $\Mat{\Sigma}^+$ and $\Mat{\Sigma}^-$, \ie, the similarity metrics obtained from positive and negative pairs in one domain, to make compact clusters of triplets that belong to another domain. This, results to better alignment of the two domains. Here, colors indicate domains.}
\label{fig1:teaser}
\vspace{-4mm}
\end{figure}

In this paper, we introduce an approach to improve the problem of face recognition from unseen domains by learning semantically meaningful representations. To this end, we are motivated by recent works in few-shot learning~\cite{li2019distribution}, domain generalization~\cite{dou2019domain} and face recognition~\cite{huang2020improving}, revealing a general fact that, in training a model, it is beneficial to exploit notions of semantic consistency between training data coming from various sources. Therefore, we introduce \CDT (CDT) loss based on the triplet objective~\cite{schroff2015facenet}, that learns useful features by considering two domains, where the similarity metrics provided by one domain are utilized in another domain to learn compact feature clusters of identities (Fig~\ref{fig1:teaser}).

Such similarity metrics are encoded by means of covariance matrices, borrowing the idea from~\cite{li2019distribution}.
Different from ~\cite{li2019distribution}, however, instead of using class-specific covariance matrices, we cast the problem in domain alignment regime, where our model first estimates feature distributions between the anchor and positive/negative samples, namely the similarity metrics derived from positive/negative pairs in one domain (\ie, $\Mat{\Sigma}^+$ and $\Mat{\Sigma}^-$ in Fig~\ref{fig1:teaser}).
%
%
Then, we utilize these similarity metrics and apply them to triplets of another domain to learn compact clusters. As supported by theoretical insights and experimental evaluations, our CDT loss aligns distributions of two domains in a discriminative manner. Furthermore, by leveraging a meta-learning framework, our network parameters are further enforced to learn generalized features under domain shift, following recent studies~\cite{dou2019domain, guo2020learning}. 


Our experiments demonstrate the effectiveness of our approach equipped with the \CDT loss, consistently outperforming the state-of-the-art on practical scenarios of face recognition for unknown ethnicity using the Cross-Ethnicity Faces (CEF)~\cite{sohn_iclr_2019} and Racial Faces in-the-Wild (RFW)~\cite{wang2019racial} benchmark datasets. Furthermore, it can satisfactorily handle face recognition across other variations as shown by empirical evaluations. 


To summarize, we introduce an effective \CDT loss function which utilizes explicit similarity metrics existing in one domain, to learn compact clusters of identities from another domain. This, results to learning semantically meaningful representations for face recognition from unseen domains. 
To further expose the network parameters to domain shift, under which more generalized features are obtained, we also incorporate the new loss in a model-agnostic learning pipeline. Our experiments show that our proposed method achieves state-of-the-art results on the standard face recognition from unseen domain datasets. 

%% file: 2_related.tex
\section{Related Work}
\paragraph{Face Recognition.} 
Following the success of deep neural networks, recent works on face recognition have tremendously improved the performances~\cite{yi2014learning,schroff2015facenet,peng_iccv2017,wang2018cosface,deng2019arcface,shi2020towards,aruni_eccv2020}, thanks to large amount of labeled data being at the disposal. Many loss designs are shown to be effective in large scale network training, \ie, CosFace~\cite{wang2018cosface} proposes to squeeze the classification boundary with a loss margin in cosine manifold, ArcFace~\cite{deng2019arcface} combines the boundary margin with an angular margin to have better classification effect. Very recently, URFace~\cite{shi2020towards} proposes sample-level confidence weighted cosine loss with an adversarial de-correlation loss to achieve better feature representations.



Generally, face recognition algorithms conjecture that the training data (e.g. CASIA-WebFace~\cite{yi2014learning} or MS-Celeb-1M~\cite{guo2016ms}) and testing data follow similar distributions. Recent works~\cite{guo2020learning,sohn_iclr_2019} however, demonstrate unsatisfactory performance of such systems due to their poor generalization ability to handle unseen data in practice. This, makes face recognition models that are aware of test-time distribution changes, be more favorable. Furthermore, in minimizing distribution discrepancy, it is crucial to consider class information to avoid mis-alignment of samples from different categories in different domains~\cite{kang2019contrastive}.
\vspace{-3mm}
\paragraph{Meta-learning and Domain Generalization.}
It is now an accepted fact that meta-learning (a.k.a. learning to learn~\cite{thrun2012learning}) can boost generalization ability of a model. The episodic training scheme originated from Model-Agnostic Meta-Learning (MAML)~\cite{finn2017model} has been widely used to address few shot learning~\cite{sung2018learning,snell2017prototypical}, domain generalization for object recognition from unseen distributions~\cite{li2018domain, balaji2018metareg, li2019feature} and very recently face recognition from unseen domains~\cite{guo2020learning}. The underlying idea is to simulate the train/test distribution gap in each round of training through creating episodic train/test splits out of the available training data. Some other examples include Model-Agnostic learning of Semantic Features (MASF)~\cite{dou2019domain} which aligns a soft confusion matrix to retain knowledge about inter-class relationships, Feature Critic Networks~\cite{li2019feature} which proposes learning an auxiliary loss to help generalization and Meta-Learning Domain Generalization (MLDG)~\cite{li2017learning} which generates domain shift during training by synthesizing virtual domains within each batch. 

Efforts in addressing unseen scenarios are either on transferring existing class variances to under-represented classes~\cite{yin_cvpr2019}, or learning universal features through various augmentations~\cite{shi2020towards}. A recent effort is Meta Face Recognition (MFR)~\cite{guo2020learning}, in which a loss is composed of distances of hard samples, identity classification and the distances between domain centers. 
However, simply enforcing alignment of the centers of training domains does not necessarily align their distributions and may lead to undesirable effects, \eg, aligning different class samples from different domains~\cite{kang2019contrastive}. As a result, this loss component does not always improve recognition (see w/o da. rows in Asian and Caucasian sections of Table 10 in~\cite{guo2020learning}).


%% file: 3_method.tex
\section{Proposed Method}
\label{sec3:Proposed}
In this section, we present our approach to improve the problem of face recognition from unseen domains by learning semantically meaningful representations. To do so, we are inspired by recent works~\cite{li2019distribution,dou2019domain,huang2020improving}, showing that, in training a model, it is beneficial to exploit notions of semantic consistency between data coming from different distributions. We learn semantically meaningful features by enforcing compact clusters of identities from one domain, where the compactness is measured by underlying similarity metrics that belong to another domain with different statistics. In fact, we distill the knowledge encoded as similarity metrics across the domains with different label spaces. 


We start with introducing the overall network architecture. Our architecture closely follows a typical image/face recognition design. It consists of a representation-learning network $f_r(\cdot~;\theta_r)$, parametrized by $\theta_r$, an embedding network $f_e(\cdot~;\theta_e)$, parametrized by $\theta_e$ and a classifier network $f_c(\cdot~;\theta_c)$, parametrized by $\theta_c$. 
Following the standard setting~\cite{balaji2018metareg,dou2019domain, guo2020learning}, both $f_c(\cdot~;\theta_c)$ and $f_e(\cdot~;\theta_e)$ are light networks, a couple of Fully Connected (FC) layers, which take inputs from $f_r(\cdot~;\theta_r)$. More specifically, forwarding an image $I$ through $f_r(\cdot)$ outputs a tensor $f_r(I) \in \mathbb{R}^{H \times W \times D}$ that, after being flattened, acts as input to both the classifier $f_c(\cdot)$ and the embedding network $f_e(\cdot)$.


Before delving into more details, we first review some basic concepts used in our formulation. Then, we provide our main contribution to learn generalized features from multiple source domains with disjoint label spaces. Finally, we incorporate the solution into a model-agnostic algorithm, originally based on Model-Agnostic Meta-Learning (MAML)~\cite{finn2017model}. 

\subsection{Notation and Preliminaries} 
\label{sec:notation}
Throughout the paper, we use  bold lower-case letters (\eg, $\Vec{x}$) to show column vectors and bold upper-case letters (\eg, $\Mat{X}$) for matrices. The $d \times d$ identity matrix is denoted by $\Mat{I}_d$. By a {\it tensor} $\Ten{X}$, we mean a multi-dimensional array of order $k$ , \ie, $\Ten{X} \in \mathbb{R}^{d_1 \times \cdots \times d_k}$. $[\Ten{X}]_{i, j, \cdots,k}$ denotes the element at position $\{i, j, \cdots,k\}$ in $\Ten{X}$.

In Riemannian geometry, the Euclidean space $\mathbb{R}^{d}$ is a Riemannian manifold equipped with the inner product defined as $\left \langle \Vec{x} , \Vec{y} \right \rangle = \Vec{x}^T \Mat{\Sigma} ~\Vec{y}, \Vec{x}, \Vec{y} \in \mathbb{R}^{d}$~\cite{faraki2015more,faraki2018comprehensive}. The class of Mahalanobis distances in $\mathbb{R}^{d}$, $d: \mathbb{R}^d \times \mathbb{R}^d \rightarrow \mathbb{R}^+$, is denoted by
\begin{equation}
d_{\Mat{\Sigma}}(\Vec{x} ,\Vec{y}) = \sqrt{(\Vec{x} - \Vec{y})^T \Mat{\Sigma}~ (\Vec{x} - \Vec{y})} \;,
	\label{eqn6:Mahal_dist}
\end{equation}
\noindent
where $\Mat{\Sigma} \in \mathbb{R}^{d \times d}$ is a  Positive Semi-Definite (PSD) matrix~\cite{kulis2012metric}, region covariance matrix of features being one instance~\cite{faraki2015material,faraki2014fisher}.
This, boils down to the Euclidean ($l_2$) distance  when the metric matrix, $\Mat{\Sigma}$, is chosen to be $\Mat{I}_d$. The motivation behind Mahalanobis metric learning is to determine $\Mat{\Sigma}$ such that $d_\Sigma(\cdot,\cdot)$ endows certain useful properties by expanding or shrinking axes in  $\mathbb{R}^{d}$. 

In a general deep neural network for metric learning, one relies on a FC layer with weight matrix  $\Mat{W} \in \mathbb{R}^{D \times d}$ immediately before a loss layer (\eg contrastive~\cite{hadsell2006dimensionality} or triplet~\cite{schroff2015facenet}) to provide the embeddings of the data to a reduced dimension space ~\cite{faraki2017large,faraki2017no,schroff2015facenet}. Then, given the fact that $\Mat{\Sigma}$ is a PSD matrix and can be decomposed as $\Mat{\Sigma} = \Mat{W}^T\Mat{W}$, the squared $l_2$ distance between two samples $x$ and ${y}$ (of a batch) 
passing through a network is computed as 

\begin{align} \notag
d_\Sigma^2\big({x} , {y}\big) &= \| \Mat{W} \big(f({x}) - f({y}) \big) \|_2^2 \\  &= \big(f({x}) - f({y})\big)^T\Mat{\Sigma}~ \big(f({x}) - f({y})\big)   \;, \noindent
\label{eqn:deep_mahal_dist}
\end{align}
where $f(x) \in \mathbb{R}^d$ denotes functionality of the network on an image $x$. 



In this work, we use positive (negative) image pairs to denote face images with equal (different) identities. Moreover, a triplet, (anchor, positive, negative), consists of one anchor face image, another sample from the same identity and one image from a different identity.

\subsection{Cross-Domain Similarity Learning}
\label{sec:cross_domian}
Here, we tackle the face recognition scenario where during training we observe $k$ source domains, each with different attributes like ethnicity. At test time, a new target domain is presented to the network which has samples of individuals with different identities and attributes. We formulate this problem as optimizing a network using a novel loss based on the triplet loss~\cite{schroff2015facenet} objective function, which we dub \CDT loss. \CDT loss, accepts inputs from two domains $^i\mathcal{D}$ and $^j\mathcal{D}$, estimates underlying distributions of positive and negative pairs from one domain (\eg, $^j\mathcal{D}$), to measures the distances between (anchor, positive) and (anchor, negative) samples of the other domain (\eg, $^i\mathcal{D}$), respectively. Then, using the computed distances and a pre-defined margin, the standard triplet loss function is applied.

Let $^{j}\mathbb{T}=\{^{j}(a,p,n)_b\}_{b=1}^{B_j}$ represents a batch of $B_j$ triplets from the $j$-th domain, $j \in 1 \cdots k$, from which we can consider positive samples $^{j}\mathbb{I}^+=\{^{j}(a,p)_b\}_{b=1}^{B_j}$. For simplicity we drop the superscript $j$ here. We combine all local descriptors of each image to estimate the underlying distribution by a covariance matrix. Specifically, we forward each positive image pair $(a,p)$, through $f_r(\cdot)$ to obtain the feature tensor representation  $f_r(a), f_r(p) \in \mathbb{R}^{H \times W \times D}$. We cast the problem in the space of pairwise differences. Therefore, we define the tensor $\Ten{R}^+ = f_r(a) - f_r(p)$. Next, we flatten the resulting tensor $\Ten{R}^+$ into vectors $\{\Vec{r}_i^+\}_{i=1}^{HW}, \Vec{r}_i^+ \in \mathbb{R}^D$. This allows us to calculate a covariance matrix of positive pairs in pairwise difference space as  
\begin{equation}
	\Mat{\Sigma}^+ = \frac{1}{N-1} \sum_{i=1}^N \left(\Vec{r}^+_i - \Vec{\mu}^+ \right)\left(\Vec{r}^+_i - \Vec{\mu}^+ \right)^T\;,
	\label{eqn3:cov_similar}
\end{equation}
\noindent
where $N= B H W$ and $\Vec{\mu}^+ = \frac{1}{N} \sum_{i = 1}^{N} \Vec{r}^+_i$.

Similarly, using $B$ negative pairs $\mathbb{I}^-=\{(a,n)_b\}_{b=1}^{B}$, we find $\Ten{R}^- = f_r(a) - f_r(n)$ for each $(a,n)$ and flatten $\Ten{R}^-$ into vectors $\{\Vec{r}_i^-\}_{i=1}^{HW}, \Vec{r}_i^- \in \mathbb{R}^D$. This enables us to define a covariance matrix of negative pairs as 
\begin{equation}
	\Mat{\Sigma}^- = \frac{1}{N-1} \sum_{i=1}^N \left(\Vec{r}^-_i - \mu^- \right)\left(\Vec{r}^-_i - \mu^- \right)^T\;,
	\label{eqn3:cov_dissimilar}
\end{equation}
\noindent
where $N= B H W$ and $\Vec{\mu}^- = \frac{1}{N} \sum_{i = 1}^{N} \Vec{r}^-_i$.

Considering that a batch of images has adequate samples, this will make sure a valid PSD covariance matrix is obtained, since each face image has $HW$ samples in covariance computations. Furthermore, samples from a large batch-size can satisfactorily approximate domain distributions according to~\cite{andrew2013deep, dorfer2015deep}.

Our Cross Domain Triplet loss, $l_{cdt}$, works in a similar fashion by utilizing the distance function $d^2_{\Mat{\Sigma}}(\cdot,\cdot)$ defined in~\eqref{eqn:deep_mahal_dist}, to compute distance of samples using the similarity metrics from another domain. Given triplet images $^{i}\mathbb{T}$ from domain $^i\mathcal{D}$ and $^{j}\Mat{\Sigma}^+$, $^{j}\Mat{\Sigma}^-$ from domain $^j\mathcal{D}$ computed via~\eqref{eqn3:cov_similar} and~\eqref{eqn3:cov_dissimilar}, respectively, it is defined as
\vspace{-3mm}

\begin{align}
l&_{cdt}\big(^{i}\mathbb{T}, ^{j}\mathbb{T}; \theta_r  \big) = \\ 
&\frac{1}{B}\sum_{b=1}^{B} \Big[\frac{1}{HW} \sum_{h=1}^{H} \sum_{w=1}^{W} d^2_{^{j}\Mat{\Sigma}^+}([f_r(^{i}{a}_b)]_{h,w},[f_r(^{i}{p}_b)]_{h,w}) -  \notag\\  
&\frac{1}{HW} \sum_{h=1}^{H} \sum_{w=1}^{W} d^2_{^{j}\Mat{\Sigma}^-}([f_r(^{i}{a}_b)]_{h,w},[f_r(^{i}{n}_b)]_{h,w})  + \tau \Big]_+    \notag
\end{align}
where $\tau$ is a pre-defined margin and $[\cdot]_+$ is the hinge function. We utilize class balanced sampling to provide inputs to both covariance and \CDT loss calculations as this has been shown to be more effective in long-tailed recognition problems~\cite{guo2020learning,liu2019large}. 

\begin{figure*}[!tb]
\centering
\includegraphics[scale=0.3]{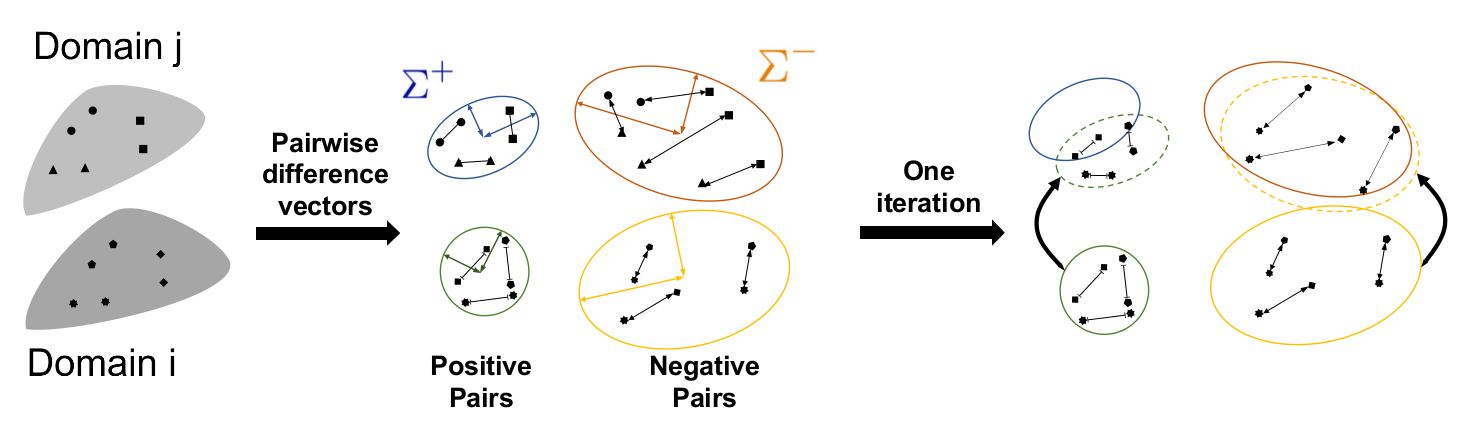}
\caption{A schematic view of our \CDT loss. In one iteration, given samples of two domains $i$ and $j$ with their associated (disjoint) labels are available, covariance matrices in difference spaces of positive and negative pairs of domain $j$ are calculated and used to make positive and negative pairs of domain $i$, close and far away, respectively. This, makes compact clusters of identities while aligning the distributions. Note that alignments of positive and negative pairs are done simultaneously in a unified manner  based on the triplet loss.}
\label{fig:diagram}
\vspace{-4mm}
\end{figure*}

\paragraph{Insights Behind our Method.}
Central to our proposal, is distance of the form $\Vec{r}^T \Mat{\Sigma}~ \Vec{r}$, defined on samples of two domains with different distributions. If $\Vec{r}$ is drawn from a distribution then the multiplications with $\Mat{\Sigma}$ results in a distance according to the empirical covariance matrix, which optimizing over the entire points translates to alignment of the domains. More specifically, assuming that $\Mat{\Sigma}$ is PSD, then eigendecomposition exists, \ie, $\Mat{\Sigma} = \Mat{V} \Mat{\Lambda} \Mat{V}^T$. Expanding the term leads to
\vspace{-2mm}

\begin{align} \notag
\Vec{r}^T \Mat{\Sigma} \Vec{r} &= \big( \Mat{\Lambda}^\frac{1}{2} \Mat{V}^T \Vec{r} \big)^T  \big( \Mat{\Lambda}^\frac{1}{2} \Mat{V}^T \Vec{r} \big)  =  \big\|  \Mat{\Lambda}^\frac{1}{2} \Mat{V}^T \Vec{r} \big\|^2_2 
\label{eqn:intuition}
\end{align}

which correlates $\Vec{r}$ with the eigenvectors of $\Mat{\Sigma}$ weighted by the corresponding eigenvalues. This, attains its maximum when $\Vec{r}$ is in the direction of leading eigenvectors of the empirical covariance matrix $\Mat{\Sigma}$. In other words, as the eigenvectors of $\Mat{\Sigma}$ are directions where its input data has maximal variance, minimizing this term over $\Vec{r}$ vectors results to alignment of the two data sources. Fig~\ref{fig:diagram} depicts the underlying process in our loss.

\subsection{A Solution in a Model Agnostic Framework}
\label{sec:maml_ours}
Following recent trends in domain generalization tasks, we employ gradient-based meta-train/meta-test episodes under a model-agnostic learning framework to further expose the optimization process to distribution shift~\cite{dou2019domain, li2019feature, guo2020learning}. Algorithm ~\eqref{alg:alg1} summarizes our overall training process. More specifically, in each round of training, we split input source domains into one meta-test and the remaining meta-train domains. We randomly sample $B$ triplets from each domain to calculate our losses. First, we calculate two covaraince matrices, $\Mat{\Sigma}^+$ and $\Mat{\Sigma}^-$, as well as temporary set of parameters, $\Theta'$, based on the summation of a classification and triplet losses, $L_s$. The network is trained to semantically perform well on the held-out meta-test domain, hence $\Mat{\Sigma}^+$, $\Mat{\Sigma}^-$ and $\Theta'$ are used to compute the loss on the meta-test domain, $L_t$. This loss has additional CDT loss, $l_{cdt}$, to also involve cross-domain similarity for domain alignment. In the end, the model parameters are updated by accumulated gradients of both $L_s$ and $L_t$, as this has been shown to be more effective than the original MAML~\cite{antoniou2018train, guo2020learning}. Here, the accumulated $L_t$ loss provides extra regularization to update the model with higher-order gradients. In the following, we provide details of the identity classification and the triplet losses. 

Having a classification training signal is crucial to face recognition applications. Hence, we use the standard Large Margin Cosine Loss (LMCL)~\cite{wang2018cosface} as our identity classification loss which is as follows 
\vspace{-2mm}
\begin{align}
	l&_{cls}\big(I_i; \theta_r,\theta_c  \big) = \\ \notag
	&-log~ \frac{\exp{(s~\Vec{w}_{y_i}^T f_c(I_i) - m)}}  {\exp{(s~\Vec{w}_{y_i}^T f_c(I_i) - m)} + \sum_{y_j \neq y_i} \exp{(s~\Vec{w}_{y_j}^T f_c(I_i))}} \;,
	\label{eqn3:lmcl}
\end{align}
\noindent
where $y_i$ is the ground-truth identity of the image $I_i$, ${f}_c(\cdot)$ is the classifier network, $\Vec{w}_{y_i}$ is the weight vector of the identity $y_i$ in $\theta_c$, $s$ is an scaling multiplier and $m$ is the margin.

We further encourage $f_r$ network to learn locally compact semantic features according to identities from one domain. To this end, we use the triplet loss. Using the standard $l_2$ distance function $\|\cdot\|_2$, the triplet loss function provides training signal such that for each triplet, the distance between $a$ and $n$ becomes greater than the distance between $a$ and $p$ plus a predefined margin $\rho$. More formally,   


\vspace{-6mm}
\begin{align}
l_{trp}&\big(\mathbb{T}; \theta_r,\theta_e  \big) = \\ \notag
&\frac{1}{B}\sum_{b=1}^{B}\Big[ \big\|f_e(a_b) - f_e(p_b)\big\|_2^2 - \big\|f_e(a_b) - f_e(n_b)\big\|_2^2  + \rho \Big]_+ \;.
\label{eqn:triplet_loss}
\end{align}
Note that $\Vec{w}_{y}$, ${f_c}(I)$ and ${f_e}(I)$ are $l_2$ normalized prior to computing the loss. Furthermore, $f_c(\cdot)$ and $f_e(\cdot)$ operate on the extracted representation by $f_r(\cdot)$. 

\vspace{-2mm}

\begin{algorithm}[!tb]
\caption{: Learning Generalized Features for Face Recognition.}
\label{alg:alg1}
\begin{algorithmic}[1]
\small
\REQUIRE
~\\
Source domains $\mathcal{D} = \big[^1\mathcal{D}, ^2\mathcal{D}_2, \cdots, ^k\mathcal{D}]$; \\
Batch-size $B$; \\
Hyper-parameters $\alpha,\beta,\lambda$; \\
\vspace{0.1cm}
\ENSURE
~\\
Learned parameters: $\hat{\Theta} = \{\hat{\theta}_r,\hat{\theta}_c,\hat{\theta}_e\}$
\vspace{0.1cm}
 \noindent
 \STATE Initialize parameters $\Theta = \{\theta_r,\theta_c,\theta_e\}$
 \REPEAT
 \STATE Initialize the gradient accumulator: $G_\Theta \leftarrow 0$ 
 \FOR{ each $^i\mathcal{D}$ (meta-test domain) in $\mathcal{D}$ }
    \FOR{ each $^j\mathcal{D}$, $i \neq j$ (meta-train domain) in $\mathcal{D}$}
        \STATE Sample $B$ triplets $^i\mathbb{T}$, from $B$ identities of $^i\mathcal{D}$
        \STATE Sample $B$ triplets $^j\mathbb{T}$, from $B$ identities of $^j\mathcal{D}$
        \STATE Compute $L_{s} \leftarrow \mathbb{E}[l_{cls}(^j\mathbb{T};\theta_r,\theta_c)] + l_{trp}(^j\mathbb{T};\theta_r,\theta_e)$
        \STATE Compute $\Theta' \leftarrow \Theta -\alpha \nabla_{(\Theta)} L_s$
        \STATE Compute $^j\Mat{\Sigma}^+$ and $^j\Mat{\Sigma}^-$ using positive and negative pairs of $^j\mathbb{T}$
        \STATE Compute $L_{t} \leftarrow \mathbb{E}[l_{cls}(^i\mathbb{T};\theta'_r,\theta'_c)] + l_{trp}(^i\mathbb{T};\theta'_r,\theta'_e) + l_{cdt}(^i\mathbb{T}, ^j\mathbb{T};\theta'_r)$
    \ENDFOR 
 \STATE $G_\Theta \leftarrow G_\Theta + \lambda \nabla_\Theta L_s + (1-\lambda) \nabla_\Theta L_t$ 
 \ENDFOR
 \STATE Update model parameters: $\Theta  \leftarrow \Theta  - \frac{\beta}{k} G_{\Theta}$
 \UNTIL{convergence}
\end{algorithmic}
\end{algorithm}


        

 

%% file: 4_exp.tex
\section{Experiments}
\label{sec4:Expr}
In this part, we first provide our implementation details for the sake of reproducibility. Then, we provide experimental results for face recognition from unseen domains. We conclude this section by ablation analysis of important components of our algorithm. To the best of our knowledge, Meta Face Recognition (MFR)~\cite{guo2020learning} is the very recent work that addresses the same problem. Therefore, we compare our method against MFR in all evaluations. We also consider the performances of CosFace~\cite{wang2018cosface}, Arcface~\cite{deng2019arcface}    and URFace~\cite{shi2020towards}  as our baseline.
\vspace{-4mm}
\paragraph{Implementation Details.}Our implementation was done in PyTorch~\cite{paszke2017automatic}. For our baseline model, we utilized a 28-layer ResNet with a final FC layer that generates an embedding space of $\mathbb{R}^{256}$, \ie, the final generalized feature space at test time. In our design, $f_r(\cdot;\theta_r)$ is the backbone immediately before this FC layer. $f_c(\cdot)$ generates logits immediately after it, while $f_e(\cdot)$ stacks an additional FC layer, to map inputs to a lower dimensional space of $\mathbb{R}^{128}$. As for the optimizer, we adopted stochastic gradient descent with the weight decay of $0.0005$ and momentum of $0.9$. In Algorithm~\eqref{alg:alg1}, the batch-size $B$ is set to $128$, $\alpha$ and $\beta$ to $1e^{-4}$ decaying to half every 1K steps and $\lambda$ to $0.7$. We set the classifier margin $m$ to $0.5$ and both  margins $\tau$ and $\rho$ to $1$.

\subsection{Cross Ethnicity Face Verification/Identification}
\label{sec4:CrossEth}
As our first set of experiments, we tackled the problem of face recognition from unseen ethnicity. Here, the evaluation protocol is leave-one-ethnicity-out, \ie, excluding samples of one ethnicity (domain) from training and then evaluating the performances on the samples of the held-out domain. 

\paragraph{Datasets.}
For evaluation, there exist two recent datasets recommended for studying facial cross ethnicity recognition performances, namely the Cross-Ethnicity Faces (CEF)~\cite{sohn_iclr_2019} and Racial Faces in-the-Wild (RFW)~\cite{wang2019racial} datasets. CEF dataset is selected from MS-Celeb-1M~\cite{guo2016ms}, consisting of four types of ethnicity images, Caucasian, African-American, East-Asian and South-Asian. We combined the last two sets into one ethnicity domain, Asian. There are 200 identities with 10 different images from each ethnicity. Note that, the identities in CEF are disjoint from the MS-Celeb-1M dataset, which we use to train our model. Similarly, RFW is another cross-ethnicity evaluation benchmark made from the MS-Celeb-1M dataset with four ethnicity subsets: Caucasian, African, Asian and Indian. CEF has only test samples, while RFW provides both training and testing data across the domains. 

\begin{table}[!t]
\centering 
\small
\begin{tabular}{cllll@{ }lll}
\hline
\multirow{2}{*}{\begin{tabular}[c]{@{}c@{}}\footnotesize{Training} \\\footnotesize{Domain(s)}\end{tabular}} & \multicolumn{3}{c}{Verification} &  & \multicolumn{3}{c}{Identification} \\ \cline{2-8} 
 & \multicolumn{1}{c}{CA} & \multicolumn{1}{c}{AA} & \multicolumn{1}{c}{AS} &  & \multicolumn{1}{c}{CA} & \multicolumn{1}{c}{AA} & \multicolumn{1}{c}{AS} \\ \hline
CA & \textbf{97.8} & 92.0 & 93.2 & & ~ \textbf{90.0} & 69.2 & 75.7 \\ \hline
AA & 91.5 & \textbf{96.4} & 92.0 &  & ~ 60.4 & \textbf{83.8} & 62.5 \\ \hline
AS & 93.3 & 91.8 & \textbf{95.6} &  & ~ 63.9 & 64.5 & \textbf{84.1} \\ \hline
All & \textbf{98.4} & \textbf{97.1} & \textbf{97.0} &  &~  \textbf{90.2} & \textbf{84.0} & \textbf{84.4} \\ \hline
\end{tabular}
\vspace{1mm}
\caption{Verification and Identification accuracy numbers in \% on the CEF~\cite{sohn_iclr_2019} testing set using our annotated MS-Celeb-1M dataset with ethnicity labels. We report the evaluation metrics when either a single domain or all domains (\ie, the last row) are used to train a CosFace model~\cite{wang2018cosface}. Testing data is either from CA: Caucasian, AA: African-American or AS: Asian domain, from disjoint identities to those of the training data. As shown in the results, regardless of the amount of data, a model trained on samples from a specific domain, performs much better on the testing data from the same domain. This, suggests the importance of having bias free training data when evaluating generalized face recognition algorithms. The top two results are highlighted in bold.} 
\vspace{-4mm}
\label{tab1:cosface_single_domain}
\end{table}

\paragraph{Effect of Ethnicity Bias in Training Data.}
Note that, most public face datasets are collected from the web by querying celebrities. This, leads to significant performance bias towards the Caucasian ethnicity. For example, 82\% of the data are Caucasian images in the MS-Celeb-1M dataset, while there are only 9.7\% African-American, 6.4\% East-Asian and less than 2\% Latino and South-Asian combined altogether~\cite{sohn_iclr_2019}.

To further highlight the influence of having ethnicity bias in training data of face recognition models, we performed an experiment using our annotated MS-Celeb-1M dataset with ethnicity labels and the CEF test set. To train a model, we considered two cases either 1) only with training samples of a single test ethnicity or 2) all training samples from all of the ethnicities. We report two standard evaluation metrics, verification accuracy and identification accuracy. To calculate the verification accuracy, we follow the standard protocol suggested by~\cite{huang2008labeled}, from which 10 splits are constructed and the overall average number is reported. Each split contains 900 positive and 900 negative pairs and the accuracy on each split is computed using the threshold found from the remaining 9 splits. As suggested by~\cite{sohn_iclr_2019}, in facial recognition systems, the drop in the identification accuracy is larger when dealing with face images from a novel ethnicity during test time. Therefore, we also report identification accuracy numbers suggested by~\cite{sohn_iclr_2019}. More specifically, we consider a positive pair of reference and query to be correct, if there is no other image from a different identity that is closer in distance to the reference, than the query.

We show the results of this experiment in Table~\ref{tab1:cosface_single_domain}. Several conclusions can be drawn here. First, the results indicate that a network trained only on samples of a specific ethnicity, tends to perform highly better on the same ethnicity, regardless of the amount of training samples. Second, the second best numbers in each column are very close to their corresponding upper bound shown in the last row ("All" here means when all training samples from all of the ethnicities are considered during training). Comparatively, the gaps between the seen domain and unseen test domains under the identification score are very larger, exceeding 20\% between CA and AA when only Caucasian samples are used during training. This, clearly shows that such a bias may invalidate conclusions drawn from performance of facial recognition systems in unseen domains. A related study~\cite{wang2018devil} shows the impact of noise in training data in face recognition.

We note that, for pre-training, the method of MFR utilizes the MS-Celeb-1M dataset, but without removing target ethnicity samples from training data. As we have experimentally observed and discussed above, this, causes difficulties when addressing generalized face recognition problem. Therefore, in this section, we adapt MFR, when considering face recognition from unseen domains.
Thus, we first aim to remove such a bias from our training data. To this end, we train an ethnicity classifier network on manually annotated face images with ground-truth ethnicity label. The number of images per ethnicity varies between 5K to 8K. We then, make use of the network to find correct ethnicity label of each individual in the dataset. This allow us to remove all known samples of an specific domain when addressing cross-ethnicity face recognition. 

\begin{table}[!tb]
\centering
\small
\begin{tabular}{l|l|lll|c}
\hline
\multicolumn{1}{c|}{\multirow{2}{*}{\begin{tabular}[c]{@{}c@{}}Unseen \\ Ethnicity\end{tabular}}} & \multicolumn{1}{c|}{\multirow{2}{*}{Method}} & \multicolumn{3}{c|}{TAR@FAR's of}                                                 & Rank@          \\ \cline{3-6} 
\multicolumn{1}{c|}{}                                                                             & \multicolumn{1}{c|}{}                        & \multicolumn{1}{c}{0.001} & \multicolumn{1}{c}{0.01} & \multicolumn{1}{c|}{0.1}   & 1              \\ \hline
\multirow{3}{*}{Caucasian}                                                                        
& CosFace~\cite{wang2018cosface}                      & \multicolumn{1}{c}{63.94} & 77.26                    & 91.44                      & 91.05          \\
& Arcface~\cite{deng2019arcface}       &  63.98    &   77.43   &   91.92    &   91.15  \\
& URFace~\cite{shi2020towards}   &  64.10    &   77.81   &   92.27    &   91.76  \\
& MFR~\cite{guo2020learning}                          & 64.31                     & 79.89                    & \multicolumn{1}{c|}{93.44} & 92.90          \\
& Ours                                         & \textbf{65.74}            & \textbf{82.48}           & \textbf{95.18}             & \textbf{94.25} \\ \hline
\multirow{3}{*}{\begin{tabular}[c]{@{}l@{}}African-\\ American\end{tabular}}                       
& CosFace~\cite{wang2018cosface}                                      & 89.48                     & 92.75                    & 95.12                      & 95.02          \\
& Arcface~\cite{deng2019arcface}       &  89.69    &    92.31  &    95.55   &  95.66   \\
& URFace~\cite{shi2020towards}   &  89.75    &   92.76   &    96.23   &    95.70 \\
& MFR~\cite{guo2020learning}                                          & 90.66                     & 95.01                    & 97.69                      & 97.07          \\
                                                                                                  & Ours                                         & \textbf{94.89}            & \textbf{96.98}           & \textbf{98.17}             & \textbf{97.23} \\ \hline
\multirow{3}{*}{Asian}                                                                            
& CosFace~\cite{wang2018cosface}                                      & 81.98                     & 91.63                    & 95.41                      & 93.94          \\
& Arcface~\cite{deng2019arcface}       &  82.96    &    91.81  &    96.09   & 93.92    \\
& URFace~\cite{shi2020towards}   &  84.33    &  92.44    &    96.10   &    94.37 \\
& MFR~\cite{guo2020learning}                                          & 85.54                     & 94.73                    & 96.86                      & 95.41          \\
                                                                                                  & Ours                                         & \textbf{88.74}            & \textbf{96.41}           & \textbf{98.56}             & \textbf{97.27} \\ \hline
\multirow{3}{*}{Indian}                                                                           
& CosFace~\cite{wang2018cosface}                                      & 80.54                     & 87.39                    & 94.55                      & 93.24          \\
& Arcface~\cite{deng2019arcface}       & 82.17     &    88.58  &    95.24   & 93.41    \\
& URFace~\cite{shi2020towards}   &  83.32    &    89.11  &    95.30   &    93.53 \\
& MFR~\cite{guo2020learning}                                          & 84.17                     & 89.42                    & 96.08                      & 93.94          \\
                                                                                                  & Ours                                         & \textbf{87.84}            & \textbf{92.08}           & \textbf{97.12}             & \textbf{95.18} \\ \hline
                                                                                              
\end{tabular}
\vspace{0.5mm}
\caption{Comparative results of our method against the state of the art on the four leave-one-ethnicity-out scenarios on the CEF test set~\cite{sohn_iclr_2019}. Note that our method consistently outperforms the competitors.} 
\vspace{-3mm}
\label{tab2:cef}
\end{table}

\paragraph{Training Data.}
To train our model, we used our MS-Celeb-1M dataset annotated with ethnicity labels. In the case of RFW experiments, we further trained the model using RFW train set, while following the leave-one-ethnicity-out testing protocol. Where there is only one training sample per identity, we made use of random augmentation of the image, to generate a positive pair. The augmentation is a random combination of Gaussian blur and occlusion suggested by~\cite{shi2020towards}. Since RFW has overlapping identities with MS-Celeb-1M, following MFR, we first removed samples of such identities and made MS-C-w/o-RFW, \ie, MS-Celeb-without-RFW. 


\begin{table}[!tb]
\centering
\small
\begin{tabular}{l|l|lll}
\hline
\multicolumn{1}{c|}{\multirow{2}{*}{\begin{tabular}[c]{@{}c@{}}Unseen \\ Ethnicity\end{tabular}}} & \multicolumn{1}{c|}{\multirow{2}{*}{Method}} & \multicolumn{3}{c}{TAR@FAR's of}                                                \\ \cline{3-5} 
\multicolumn{1}{c|}{}                                                                             & \multicolumn{1}{c|}{}                        & \multicolumn{1}{c}{0.001} & \multicolumn{1}{c}{0.01} & \multicolumn{1}{c}{0.1}   \\ \hline
\multirow{3}{*}{Caucasian}   
& CosFace~\cite{wang2018cosface}      & \multicolumn{1}{c}{61.15} & 70.74         & 85.50                     \\
& Arcface~\cite{deng2019arcface}       &  61.18    &   70.85   &    85.93        \\
& URFace~\cite{shi2020towards}   &  62.54    &  72.94    &   88.24         \\
& MFR~\cite{guo2020learning}          & 63.81                     & 76.06                    & \multicolumn{1}{c}{90.43} \\
& Ours         & \textbf{65.20}            & \textbf{78.80}           & \textbf{91.87}            \\ \hline
\multirow{3}{*}{African}                                                                          
& CosFace~\cite{wang2018cosface}                                      & 71.55                     & 83.40                    & 90.11                     \\
& Arcface~\cite{deng2019arcface}       & 71.82     &  84.07    &     92.11       \\
& URFace~\cite{shi2020towards}   &   73.93   &  86.34    &     93.26       \\
& MFR~\cite{guo2020learning}                                          & 75.31                     & 88.94                    & 93.67                     \\
& Ours                                         & \textbf{77.90}            & \textbf{91.17}           & \textbf{96.87}            \\ \hline
\multirow{3}{*}{Asian}                                                                            
& CosFace~\cite{wang2018cosface}                                      & 66.33                     & 80.04                    & 90.26                     \\
& Arcface~\cite{deng2019arcface}       & 66.49     &   80.97   &    91.73        \\
& URFace~\cite{shi2020towards}   &   67.55   & 81.04     &    90.89        \\
& MFR~\cite{guo2020learning}                                          & 69.02                     & 82.54                    & 91.94                     \\
& Ours                                         & \textbf{69.17}            & \textbf{82.60}           & \textbf{94.10}            \\ \hline
\multirow{3}{*}{Indian}                                                                           
& CosFace~\cite{wang2018cosface}                                      & 65.55                     & 80.17                    & 91.03                     \\
& Arcface~\cite{deng2019arcface}       &  67.39    &   81.15   &    91.44        \\
& URFace~\cite{shi2020towards}   &   69.26   &  83.03    &     91.97       \\
& MFR~\cite{guo2020learning}                                         & 71.44                     & 83.11                    & 92.22                     \\
& Ours                                         & \textbf{76.63}            & \textbf{86.70}           & \textbf{95.47}            \\ \hline
\end{tabular}
\vspace{1mm}
\caption{Comparative results of our method against the state of the art on the four leave-one-ethnicity-out scenarios of the RFW dataset~\cite{wang2019racial}. Note that our method consistently outperforms the competitors.}
\vspace{-6mm}
\label{tab3:rfw}
\end{table}

\paragraph{Evaluation Metrics.} At test time, the extracted feature vectors of each image and its flipped version are concatenated and considered as the final representation of the image. We used cosine distance to compute the distances. As for evaluating performances, we report the True Acceptance Rate (TAR) at different levels of False Acceptance Rate (FAR) such as $0.001, 0.01$ and $0.1$ using the Receiver Operating Characteristic (ROC) curve. Moreover, we report Rank-1 accuracy where a pair of probe and gallery image is considered correct if after matching the probe image to all gallery images, the top-1 result is within the same identity.

\paragraph{Results on the CEF Testing Set.} In Table~\ref{tab2:cef}, we compare our results with those of the state-of-the-art techniques across all four leave-one-domain-out scenarios. Note that there is no annotated face image as Indian in the source domains here. Our method, equipped with the \CDT loss, comfortably outperforms the recent method of MFR. 
This, we believe, clearly shows the benefits of our approach, which allows us to learn cross-domain similarities from labeled source face images, thus yielding a better representation for unseen domains.


\paragraph{Results on the RFW Dataset.} In Table~\ref{tab3:rfw}, using the RFW dataset, we compare our results with those of the state-of-the-art methods described above in the leave-one-ethnicity-out protocol. As in the previous experiment, our method is the top performing one in all cases. Under FAR@$0.1$ measure, the gap between our algorithm and the closest competitor, MFR, exceeds $3\%$ when African-American and Indian images are considered as the test domain. CosFace method is considered as the baseline and is the least performing method, indicating the importance of learning generalized features in face recognition.

\begin{table}[!tb]
\centering
\small
\begin{tabular}{l|cc|c|c|c}
\hline
\multirow{2}{*}{Method} &
  \multicolumn{2}{l|}{TAR@FAR's of} &
  \multicolumn{1}{l|}{Rank@} &
  \multicolumn{1}{l|}{\multirow{2}{*}{AUC}} &
  \multicolumn{1}{l}{\multirow{2}{*}{Acc}} \\ \cline{2-4}
 &
  \multicolumn{1}{l}{0.001} &
  \multicolumn{1}{l|}{~0.01} &
  1 &
  \multicolumn{1}{l|}{} &
  \multicolumn{1}{l}{} \\ \hline
CosFace~\cite{wang2018cosface} & 92.55          & 96.55          & 96.85         & 99.53         & 99.35          \\
LF-CNNs~\cite{wen2016latent} & -              & -              & -             & 99.3          & 98.5           \\
MFR~\cite{guo2020learning}     & 94.05          & \textbf{97.25} & \textbf{97.8} & \textbf{99.8} & \textbf{99.78} \\
Ours    & \textbf{94.20} & 97.22          & 97.75         & \textbf{99.8} & 99.76          \\ \hline
\end{tabular}
\vspace{1mm}
\caption{Comparative results of our method against the state of the art on cross-age face recognition using the CACD-VS dataset~\cite{chen2014cross}. Our method works on par with MFR.}
\label{tab4:cacd}
\vspace{-4mm}
\end{table}

\subsection{Handling Other Variations}
\label{sec4:CrossAge}
To demonstrate that our approach is generic and can handle other common variations during test time, we considered additional experiments. Following the standard protocol suggested by MFR~\cite{guo2020learning}, the full train data annotated with ethnicity is used as our source domains in all experiments.

\paragraph{Cross age face verification and identification.}
 First, we considered an experiment using the Cross-Age Celebrity Dataset (CACD-VS)~\cite{chen2014cross}. The dataset provides 4k cross-age image pairs with equal number of positive and negative pairs to form a testing subset for face verification task. The results of this experiment are shown in Table~\ref{tab4:cacd}, where we have also reported the Area Under the ROC Curve (AUC) as well as the verification accuracy numbers. The table indicates, while our method outperforms the recent method of MFR under the FAR@$0.001$ measure, works competitively very close to it under other evaluation metrics. Under the AUC measure, we obtain the exact same performance as the MFR. In all comparisons, our method outperforms the previous work of LF-CNNs~\cite{wen2016latent}.

\paragraph{Near infrared vs. visible light face recognition.}
 Next, we considered an experiment using the CASIA NIR-VIS 2.0 face database~\cite{li2013casia}. Here, the gallery face images are captured under visible light while the probe ones are under near infrared light.  Table~\ref{tab:new_nir} shows that our method consistently outperforms the competitors.
 
\begin{table}[!t]
\centering
\small
\begin{tabular}{l|lll|l}
\hline
\multicolumn{1}{c|}{\multirow{2}{*}{Method}} & \multicolumn{3}{c|}{TAR@FAR's of}                & Rank@                 \\ \cline{2-5} 
\multicolumn{1}{c|}{}                        & 0.001          & 0.01           & 0.1            & \multicolumn{1}{c}{1} \\ \hline
CosFace~\cite{wang2018cosface}     & 69.03          & 71.35          & 90.47          & 94.29                 \\
Arcface~\cite{deng2019arcface} & 69.15          & 71.62          & 91.22          & 94.36                 \\
URFace~\cite{shi2020towards}    & 70.79          & 74.67          & 93.40          & 94.72                 \\
MFR~\cite{guo2020learning}    & 72.33          & 81.92          & 95.97          & 96.92                 \\
Ours                                         & \textbf{76.41} & \textbf{84.53} & \textbf{97.68} & \textbf{97.24}        \\ \hline
\end{tabular}
\vspace{1mm}
\caption{Comparative results of our method against the state of the art on CASIA NIR-VIS 2.0 dataset~\cite{li2013casia}.}
\label{tab:new_nir}
\end{table}

\paragraph{Cross pose face recognition.}
Finally, we considered an experiment using the Multi-PIE cross pose dataset~\cite{gross2010multi,peng_iccv2017}. Similar to the previous experiment, our method outperforms the competitors as shown in Table~\ref{tab:new_MPie}.

\begin{table}[!t]
\centering
\small
\begin{tabular}{l|lll|l}
\hline
\multicolumn{1}{c|}{\multirow{2}{*}{Method}} & \multicolumn{3}{c|}{TAR@FAR's of}              & Rank@                 \\ \cline{2-5} 
\multicolumn{1}{c|}{}                        & 0.001          & 0.01           & 0.1          & \multicolumn{1}{c}{1} \\ \hline
CosFace ~\cite{wang2018cosface} & 68.49          & 98.96          & 99.92        & 99.82                 \\
ArcFace~\cite{deng2019arcface}    & 69.53          & 99.0           & 99.96        & 99.86                 \\
URFace~\cite{shi2020towards}    & 69.77          & 99.54          & 99.96        & 99.90                 \\
MFR~\cite{guo2020learning}      & 74.54          & \textbf{99.96} & \textbf{100} & \textbf{99.92}        \\
Ours                                         & \textbf{76.62} & \textbf{99.96} & \textbf{100} & \textbf{99.92}        \\ \hline
\end{tabular}
\vspace{1mm}
\caption{Comparative results of our method against the state of the art on Multi-PIE dataset~\cite{gross2010multi}.}
\label{tab:new_MPie}
\end{table}

\subsection{Ablation Analysis}
\label{sec4:Ablation}
In this section, we conduct further experiments to show the impact of important components of Algorithm~\ref{alg:alg1} on verification accuracy. For this analysis, we made use of the Caucasian subset of the RFW dataset as the unseen domain and the rest of the data from all other domains as the source domains.
 
 \vspace{1mm}
\begin{table}[!tb]
\centering
\small
\begin{tabular}{c|ccc}
\hline
\multirow{2}{*}{Method} & \multicolumn{3}{c}{TAR@FAR's of}                                              \\ \cline{2-4} 
                        & \multicolumn{1}{l}{0.001} & \multicolumn{1}{l}{~0.01} & \multicolumn{1}{l}{~~0.1} \\ \hline
w/o $l_{cls}$           & 64.24                     & 76.58                    & 89.96                   \\
w/o $l_{trp}$           & 64.65                     & 77.95                    & 91.02                   \\
w/o $l_{cdt}$           & 63.75                     & 75.90                    & 89.84                   \\
Ours (full)             & \textbf{65.20}            & \textbf{78.80}           & \textbf{91.87}          \\ \hline

\end{tabular}
\vspace{1mm}
\caption{Ablation study of the effects of different loss components of our proposal.}
\label{tab4:ablation_loss}
\vspace{-4mm}
\end{table}

We first investigate how final recognition accuracy numbers vary, when different loss components are excluded from our algorithm. Different components here are identity classification, triplet and cross-domain triplet losses, denoted by $l_{cls}$, $l_{trp}$ and $l_{cdt}$, respectively. The results of this experiment are shown in Table~\ref{tab4:ablation_loss}. As the table indicates, every component of our overall loss, contributes importantly to the final performance, as excluding either of them leads to a consistent performance drop. Further comparing across the different loss terms, we observe that our proposed CDT loss plays a more important role here, since the performances significantly drop without this loss component. For instance, FAR@$0.1$ drops by more than $2\%$ without $l_{cdt}$. 

\begin{figure}[!tb]
\centering
\includegraphics[scale=0.38]{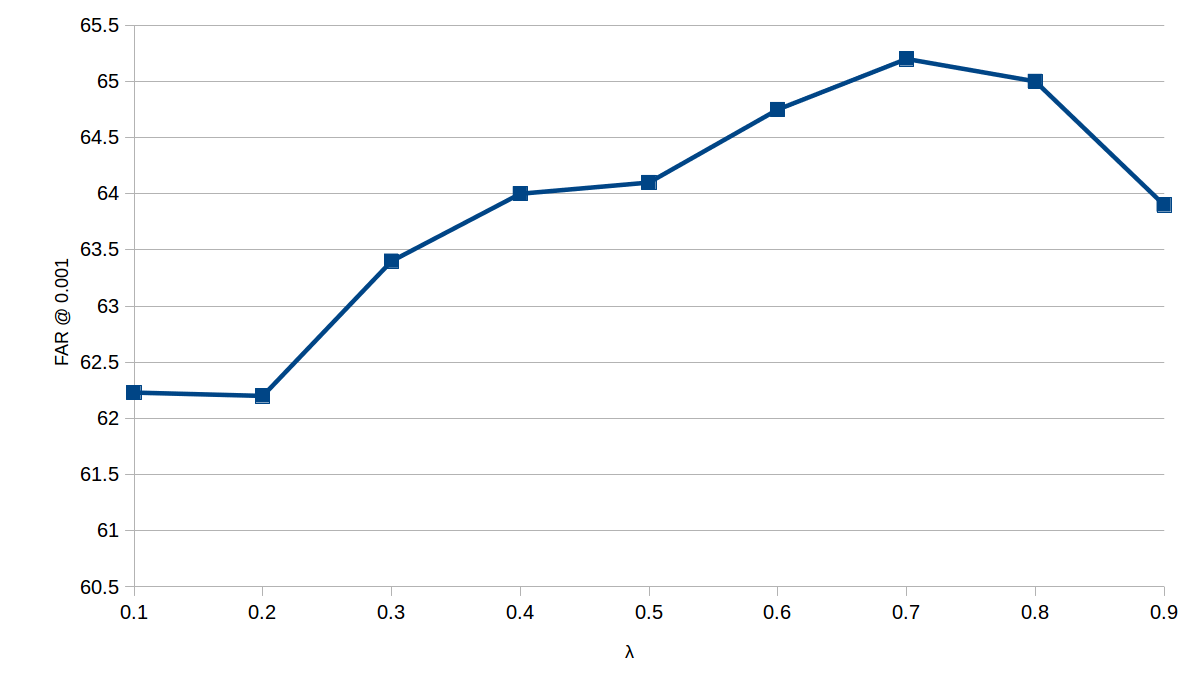}
\vspace{-1mm}
\caption{Ablation study over the hyper-parameter $\lambda$. This indicates the effect of meta-test loss which has our \CDT loss. We empirically observed that a value close to $0.7$ gives the best results.}
\label{fig:ablation}
\vspace{-4mm}
\end{figure}

Furthermore, one hyper-parameter in our approach is the contribution ratio of meta-train and meta-test losses, \ie, $L_s$ and $L_t$. This is determined by hyper-parameter $\lambda$ in Algorithm~\ref{alg:alg1}. In Fig~\ref{fig:ablation}, we show the effect of varying $\lambda$ on the FAR@$0.001$ measure, when the Caucasian subset of the RFW dataset is considered the target domain. We observe that a value close to $0.7$ gives the best results. 

